\pdfoutput=1
\documentclass[11pt]{article}
\usepackage[dvipsnames]{xcolor}

\usepackage{emnlp2021}

\usepackage{times}
\usepackage{latexsym}

\usepackage[T1]{fontenc}
\usepackage[utf8]{inputenc}

\usepackage{microtype}

\usepackage{bm}
\usepackage{amssymb}
\usepackage[tbtags]{amsmath}
\usepackage{amsfonts}

\DeclareMathOperator*{\argmax}{argmax} % no space, limits underneath in displays

\usepackage{cleveref}
\crefname{section}{\S}{}
\Crefname{figure}{Figure}{}
\Crefname{table}{Table}{}
\Crefname{algorithm}{Algorithm}{}
\Crefname{algocf}{Algorithm}{}
\Crefname{equation}{Equation}{}

\usepackage{makecell}
\usepackage{booktabs}
\usepackage{multirow}
\usepackage{tabulary}
\newcolumntype{L}[1]{>{\raggedright\let\newline\\\arraybackslash\hspace{0pt}}m{#1}}
\newcolumntype{C}[1]{>{\centering\let\newline\\\arraybackslash\hspace{0pt}}m{#1}}
\newcolumntype{R}[1]{>{\raggedleft\let\newline\\\arraybackslash\hspace{0pt}}m{#1}}
\usepackage[ruled]{algorithm2e}

\definecolor{drug}{HTML}{377EB8}
\definecolor{gene}{HTML}{4DAF4A}
\definecolor{variant}{HTML}{E41A1C}
\definecolor{relation-classification}{HTML}{0064C9}
\definecolor{argument-resolution}{HTML}{7CAB62}

\usepackage{tikz}

\usepackage{textcomp}

\newcommand{\eat}[1]{\ignorespaces}

\newcommand{\ckb}{\mbox{CKB}\xspace}
\newcommand{\ckbcore}{CKB CORE\texttrademark\xspace}
\newcommand{\ckbhard}{CKB$^{\tt HARD}$\xspace}

\newcommand{\modular}{$\tt Modular$\xspace}
\newcommand{\multiscale}{$\tt Multiscale$\xspace}
\newcommand{\allpos}{$\tt All Positive$\xspace}
\newcommand{\gain}{$\tt GAIN$\xspace}

\title{Modular Self-Supervision for Document-Level Relation Extraction}

\author{Sheng Zhang$^1$ \quad Cliff Wong$^1$ \quad Naoto Usuyama$^1$ \quad Sarthak Jain$^2$\thanks{~~Work done as an intern at Microsoft Research.} \\
    {\bf Tristan Naumann$^1$} \quad {\bf Hoifung Poon$^1$}\\
    $^1$Microsoft Research, Redmond, WA\\
    $^2$Khoury College of Computer Sciences, Northeastern University \\
    \texttt{\{shezhan,clwon,naotous,tristan,hoifung\}@microsoft.com} \\
    \texttt{jain.sar@husky.neu.edu} \\}

\begin{document}
\maketitle

\begin{abstract}

Extracting relations across large text spans has been relatively underexplored in NLP, but it is particularly important for high-value domains such as biomedicine, where obtaining high recall of the latest findings is crucial for practical applications. 
Compared to conventional information extraction confined to short text spans, document-level relation extraction faces additional challenges in both inference and learning. Given longer text spans, state-of-the-art neural architectures %such as transformers and recurrent neural networks are less effective
are less effective and task-specific self-supervision such as distant supervision becomes very noisy. 
In this paper, we propose decomposing document-level relation extraction into relation detection and argument resolution, taking inspiration from Davidsonian semantics. 
This enables us to incorporate explicit discourse modeling and leverage modular self-supervision for each sub-problem, which is less noise-prone and can be further refined end-to-end via variational EM. 
%accommodate both binary and general n-ary relation extraction.
We conduct a thorough evaluation in biomedical machine reading for precision oncology, where cross-paragraph relation mentions are prevalent. 
Our method outperforms prior state of the art, such as multi-scale learning and graph neural networks, by over 20 absolute F1 points. The gain is particularly pronounced among the most challenging relation instances whose arguments never co-occur in a paragraph.%, where our method triples the F1.
%and whose average text span is over twenty sentences across four paragraphs.

\end{abstract}
\section{Introduction}

\begin{figure}[!t]
    \centering
    \includegraphics[width=0.47\textwidth]{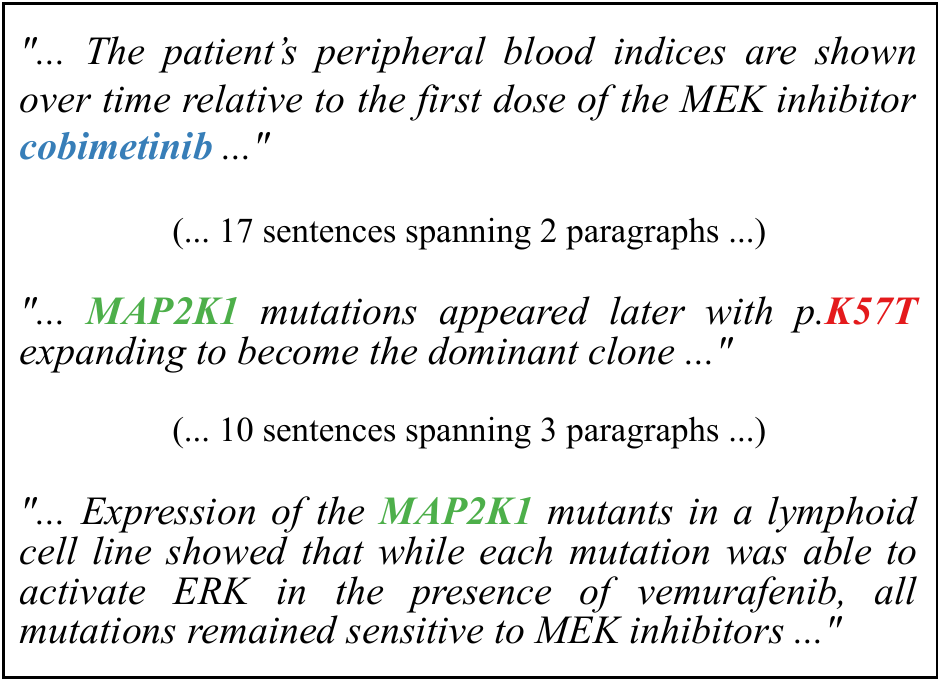}
    \caption{An example document-level drug-gene-mutation relation. %\textcolor{drug}{drug}-\textcolor{gene}{gene}-\textcolor{variant}{mutation} relations,
    \textbf{\emph{\textcolor{drug}{cobimetinib}}} never co-occurs with \textbf{\emph{\textcolor{gene}{MAP2K1}}} or \textbf{\emph{\textcolor{variant}{K57T}}} in any paragraph.}
    \label{fig:motivating-example}
\end{figure}

Prior work on information extraction tends to focus on binary relations within sentences. However, practical applications often require extracting complex relations across large text spans. 
This is especially important in high-value domains such as biomedicine, where obtaining high recall of the latest findings is crucial. 
For example, \autoref{fig:motivating-example} shows a ternary (drug, gene, mutation) relation signifying that a tumor with MAP2K1 mutation K57T is sensitive to cobimetinib, yet the entities never co-occur in any single paragraph. Such precision oncology knowledge is key for determining personalized treatment for cancer patients, but it is scattered among a vast biomedical literature of more than 30 million papers, with over 1 million being added each year\footnote{\url{http://www.ncbi.nlm.nih.gov/pubmed}}.

Recently, there has been increasing interest in cross-sentence relation extraction, but most existing work still focuses on short text spans. \citet{quirk-poon-2017-distant} and \citet{peng-etal-2017-cross} restrict extraction to three consecutive sentences and \citet{verga-etal-2018-simultaneously} to abstracts.
DocRED~\cite{yao-etal-2019-docred}, a popular document-level relation extraction dataset, consists of Wikipedia introduction sections, each with only eight sentences on average. Further, half of the relation instances reside in a single sentence, all effectively in a single paragraph.

\begin{figure*}[!t]
    \centering
    \includegraphics[width=0.97\textwidth]{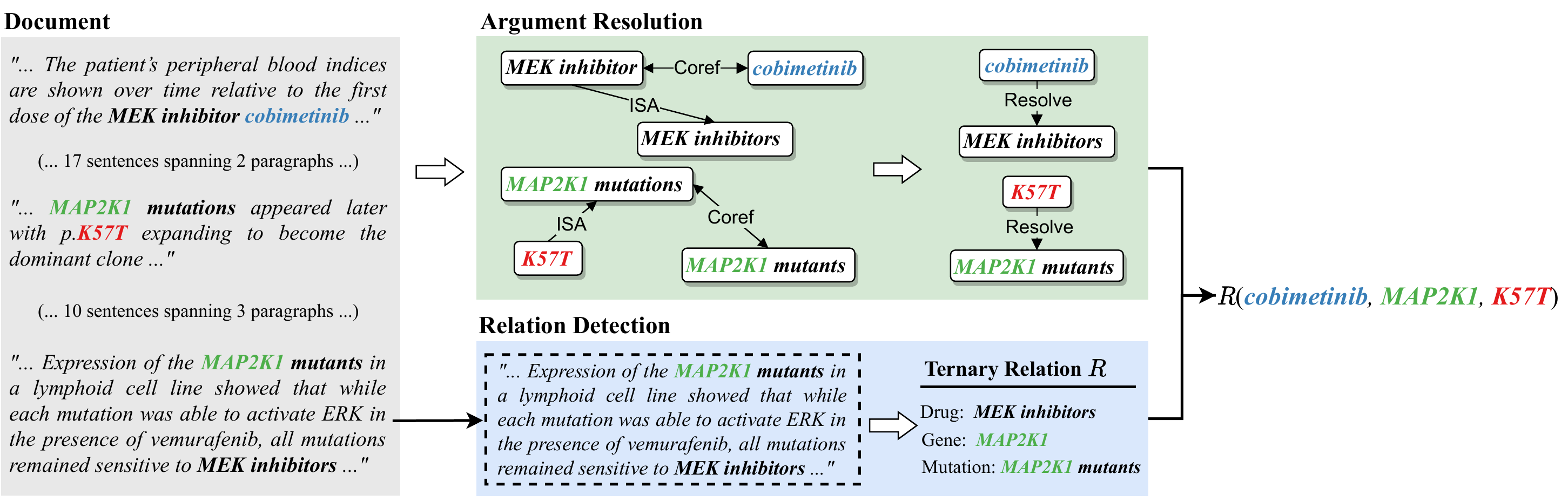}
    \caption{Decomposing document-level $n$-ary relation extraction into relation detection and argument resolution.}
    \label{fig:formulation}
\end{figure*}

To the best of our knowledge, \citet{jia-etal-2019-document} is the first to consider relation extraction in full-text articles, which is considerably more challenging. They use the CKB dataset \cite{patterson2016clinical} for evaluation where each document contains, on average, 174 sentences, spanning 39 paragraphs. 
Additionally, while prior work focuses mainly on binary relations, \citet{jia-etal-2019-document} follows \citet{peng-etal-2017-cross} to extract ternary relations. 
However, while \citet{jia-etal-2019-document} admits relation instances for which the three arguments never co-occur in a paragraph, it still requires that the two arguments for each binary subrelation co-occur at least once in a paragraph, leaving one fifth of findings out of reach for their method. 

%4 paragraphs / 21 sentences; 174 sent/39 para

All this prior work considers document-level relation extraction as a single monolithic problem, which presents major challenges in both inference and learning. Despite recent progress, there are still significant challenges in modeling long text spans using state-of-the-art neural architectures, such as LSTM and transformer. Moreover, direct supervision is scarce and task-specific self-supervision, such as distance supervision, becomes extremely noisy when applied beyond short text spans. 

In this paper, we explore an alternative paradigm by decomposing document-level relation extraction into local relation detection and global reasoning over argument resolution. 
Specifically, we represent $n$-ary relation using Davidsonian semantics and combine paragraph-level relation classification with discourse-level argument resolution using global reasoning rules (e.g., transitivity over argument resolution). Each component problem resides in short text spans and their corresponding self-supervision is much less error-prone. 
Our approach takes inspiration from modular neural networks \cite{andreas2016neural} and neural logic programming \cite{rocktaschel&riedel16} in decomposing a complex task into local neural learning and global structured integration. % (via syntactic parse in modular neural networks and backward chaining in neural logic programming). 
However, instead of learning from end-to-end direct supervision, we admit modular self-supervision for the component problems, which is more readily available. Our method can thus be viewed as applying deep probabilistic logic \cite{wang-poon-2018-deep} to combine modular self-supervision and joint inference with global reasoning rules.

%This enables us to incorporate explicit discourse modeling and leverage modular self-supervision for each sub-problem, which is less noise-prone and can be further refined end-to-end via variational EM. 
This modular approach enables us to not only handle long text spans such as full-text articles like \citet{jia-etal-2019-document}, but also expand extraction to the significant portion of cross-paragraph relations that are out of reach to all prior methods. 
We conduct a thorough evaluation in biomedical machine reading for precision oncology, where such cross-paragraph relations are especially prevalent. 
Our method outperforms prior state of the art such as multiscale learning \cite{jia-etal-2019-document} and graph neural networks~\cite{zeng-etal-2020-double} by over 20 absolute F1 points. The gain is particularly pronounced among the most challenging relations whose arguments never co-occur in a paragraph.

\section{Document-Level Relation Extraction}

Let $E_1,\dotsc,E_n$ be entities that co-occur in a document $D$. Relation extraction amounts to classifying whether a relation $R$ holds for $E_1,\dotsc,E_n$ in $D$.
For example, in \autoref{fig:formulation}, $R$ is the relation of precision cancer drug response, and $E_1, E_2, E_3$ represent drug cobimetinib, gene MAP2K1, mutation K57T, respectively. 
The relation mention spans multiple paragraphs and dozens of sentences. Direct extraction is challenging and ignores the elaborate underlying linguistic phenomena. The drug-response relation is explicitly mentioned in the last paragraph, though it is between ``\emph{MEK inhibitors}'' and ``\emph{MAP2K1 mutations}''. Meanwhile, the top paragraph states the $\tt ISA$ relation between ``\emph{cobimetinib}'' and ``\emph{MEK inhibitors}'', as apparent from the apposition. From the middle paragraph one can infer the $\tt ISA$ relation between ``\emph{K57T}'' and ``\emph{MAP2K1 mutations}''. Finally, ``\emph{MAP2K1 mutants}'' in the last paragraph can be resolved with ``\emph{MAP2K1 mutations}'' in the middle based on semantic similarity. 
Combining these, we can conclude that the drug-response relation holds for (cobimetinib, MAP2K1, K57T) in this document (\autoref{fig:formulation}), even though cobimetinib never co-occurs with MAP2K1 or K57T in any paragraph.

% Terence Parsons, Events in the Semantics of English, 1990, MIT Press
Formally, we represent $n$-ary relation extraction by neo-Davidsonian semantics \cite{parsons}:
% \vspace{-7pt}
\begin{multline*}
R_D(E_1,\cdots,E_n) \doteq \exists T\in D~\exists r.\\
[R_T(r)\wedge A_1(r,E_1)\wedge \cdots\wedge A_n(r,E_n)]% \vspace{-2pt}
\end{multline*}
Here, $T$ is a text span in $D$, $r$ is a reified event variable introduced to represent relation $R$, and the arguments are represented by binary relations with the event variable. The distributed nature of this representation makes it suitable for arbitrary $n$-ary relations and does not require drastic changes when arguments are missing or when new arguments are added. 
Given this representation, document-level relation extraction is naturally decomposed into local relation detection (e.g., classifying if $R_T(r)$ holds for some paragraph $T$) and global argument resolution (e.g., classifying $A_i(r,E_i)$). 

Entity-level argument resolution can be reduced to mention-level argument resolution $A(r,E) \doteq \exists e.~[{\tt Mention}(e,E) \wedge A(r,e)]$, where ${\tt Mention}(e,E)$ signifies that $e$ is an entity mention of $E$. Additionally, the transitivity rule applies: \[A(r,e) \wedge {\tt Resolve}(r,e,e') \implies A(r,e')\]
Here, ${\tt Resolve}(r,e,e')$ signifies that the mentions $e,e'$ are interchangeable in the context of relation mention $r$. 
For brevity, in this paper we drop the relation context $r$ and simply consider ${\tt Resolve}(e,e')$. 
If $e,e'$ are coreferent or semantically equivalent, as in (MAP2K1 mutations, MAP2K1 mutants), $\tt Resolve$ obviously holds. More generally, $\tt ISA$ (e.g., K57T and MAP2K1 mutations) and $\tt PartOf$ (e.g., a mutation and a cell line containing it) may also signify resolution:
\begin{align*}
    {\tt Coref}(e,e')\implies {\tt Resolve}(e,e')\\
    {\tt ISA}(e,e')\implies {\tt Resolve}(e,e')\\
    {\tt PartOf}(e,e')\implies {\tt Resolve}(e,e')
\end{align*}

Also, transitivity generally holds for $\tt Resolve$: 
%{\tt Resolve}(e,e')\wedge {\tt Resolve}(e',e'')\implies {\tt Resolve}(e,e'')$.
%\eat{
% \vspace{-7pt}
\begin{multline*}
{\tt Resolve}(e,e')\wedge {\tt Resolve}(e',e'')\\
\implies {\tt Resolve}(e,e'')
\end{multline*}
%}

%%%%%%%%%%%%%%%%%%%%%%%%%%%%%%%%%%%%%%%%%%%%%%%%%%%%%%%%%%
\section{Modular Self-Supervision}
\label{sec:modular}

\begin{table}[!t]
\small
\centering
\begin{tabular}{@{}ccl@{}}
\toprule
 & \multicolumn{2}{c}{\bf Modular Self-Supervision} \\ \midrule
\makecell[l]{\textbf{Relation}\\\textbf{Detection}} & \makecell[c]{Distant Supervision} & \makecell[l]{CIVC, GDKD,\\OncoKB} \\ \midrule
\multirow{2}{*}{\makecell[l]{\textbf{Argument}\\ \textbf{Resolution}}} & \makecell[c]{Data Programming} & \makecell[l]{Identical mentions,~~\\apposition} \\ \cmidrule(l){2-3}
 & \makecell[c]{Joint Inference} & Transitivity \\ \bottomrule
\end{tabular}
\caption{Modular self-supervision for relation detection and argument resolution.}
\label{tab:ssl}
\end{table}

Our problem formulation makes it natural to introduce modular self-supervision for relation detection and argument resolution (Table \ref{tab:ssl}). 

\paragraph{Relation Detection} The goal is to train a classifier for $R_T(r)$. In this paper, we consider paragraphs as candidates for $T$ and use distant supervision \cite{mintz-etal-2009-distant} for self-supervision. Specifically, knowledge bases (KBs) with known relation instances for $R$ are used to annotate examples from unlabeled text. Co-occurring mention tuples of known relations are annotated as positive examples and those not known to have relations are sampled as negative examples. These examples are then used to train a paragraph-level relation classifier. Here, we leverage the fact that paragraph-level distant supervision is much less noise-prone, but document-level relation mentions still observe similar textual patterns as paragraph-level ones, as can be seen in \autoref{fig:formulation}.

\paragraph{Argument Resolution} The goal is to train a classifier for ${\tt Resolve}(e,e')$ based on local context for entity mentions $e,e'$. As stated in the prior section, $\tt Resolve$ is strictly more general than coreference and may involve $\tt ISA$ and $\tt PartOf$ relations. For self-supervision, we introduce data programming rules that capture identical mentions and appositives.  %such as ``??'' and ``??''. 
These are used as seed self-supervision to annotate high-precision resolution instances. In turn, additional instances in the same document can be generated by applying the transitivity rule. E.g., in \autoref{fig:formulation}, by deriving ${\tt Resolve}$(cobimetinib, MEK inhibitor) in the top paragraph based on the apposition, we may annotate additional $\tt Resolve$ instances between ``\emph{cobimetinib}'' and ``\emph{MEK inhibitors}'' in the bottom paragraph. As in distant supervision, there will be noise, but on balance, such joint inference helps learn more general resolution patterns. 

%%%%%%%%%%%%%%%%%%%%%%%%%%%%%%%%%%%%%%%%%%%%%%%%%%%%%%%%
\section{Our Model}

\eat{
\begin{figure*}[!t]
    \centering
    \includegraphics[width=0.95\textwidth]{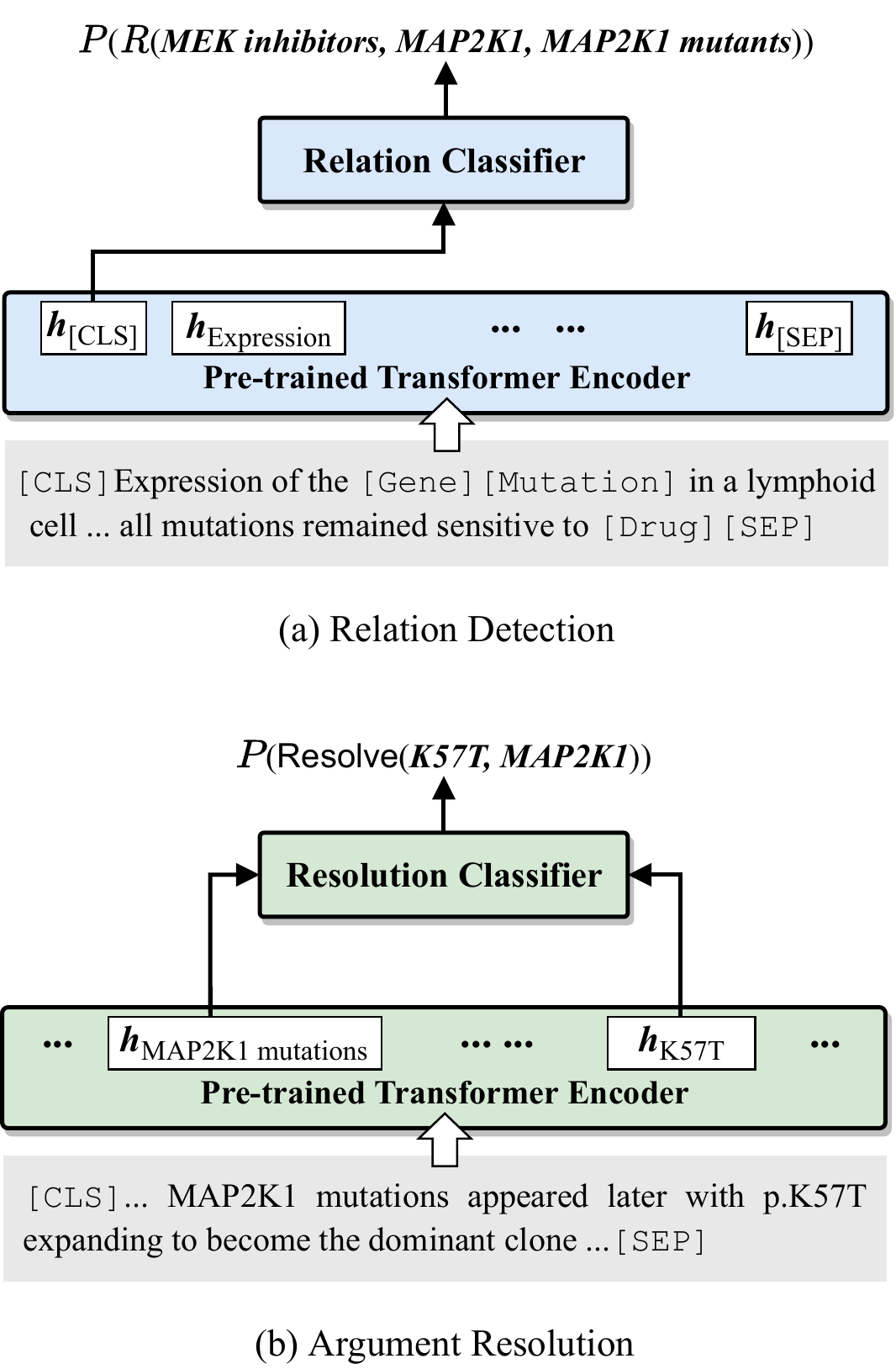}
    \caption{Neural architectures for relation detection and argument resolution.}
    \label{fig:model}
\end{figure*}
}

%{\bf TODO: Add an overall architecture of our system ~ DPL - Highlight modular self-supervision, modules - use ternary}

\begin{figure}[!t]
    \centering
    \includegraphics[width=0.43\textwidth]{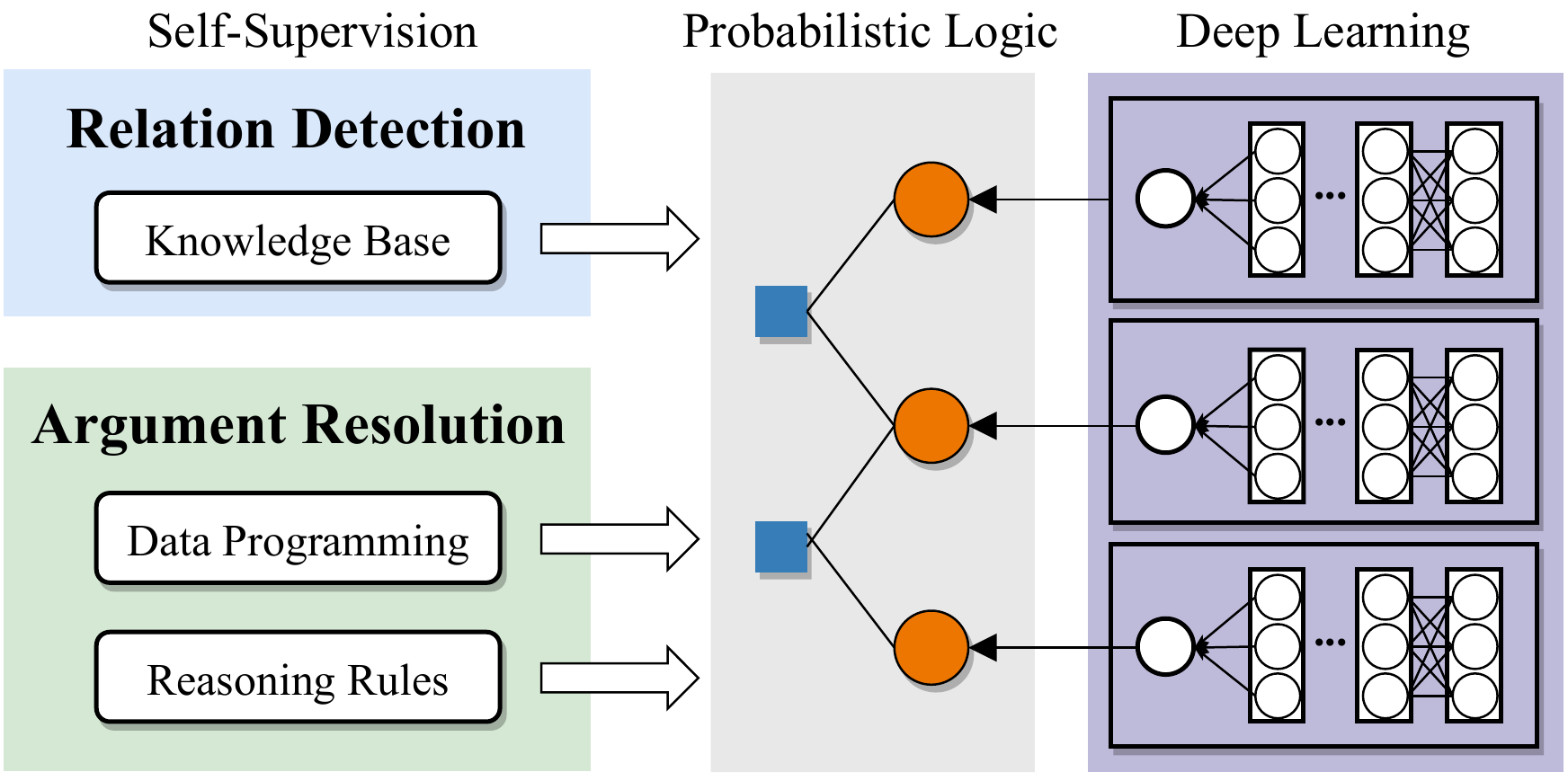}
    \caption{Our approach for document-level relation extraction applies deep probabilistic logic to incorporate modular self-supervision and joint inference for relation detection and argument resolution.}
    \label{fig:architecture}
\end{figure}

% architecture
\autoref{fig:architecture} shows our document-level relation extraction system, which uses deep probabilistic logic \cite{wang-poon-2018-deep} to incorporate modular self-supervision and joint inference. 

% prediction
\paragraph{Prediction Module} The prediction module comprises transformer-based neural models~\cite{transformer} for local relation detection and global argument resolution. %, as shown in \autoref{fig:prediction}.
For relation detection, let $(m_1,\cdots,m_n)$ be a candidate co-occurring mention tuple in a paragraph $T$. We input $T$ to a transformer-based relation classifier, with mentions $m_i$ dummified.\footnote{Alternatively, we can add entity markers for each mention.} 
For argument resolution, let $(m, m')$ be a candidate mention pair. We compute the contextual representation using a transformer model for both mentions and classify the pair using a comparison network. The input concatenates contextual representations of the entities as well as their element-wise multiplication. 
For the detailed neural architectures, see Appendix \ref{sec:neural-architecture}.

%{\bf Add details resolution = element-wise multiplication}

% supervision
\paragraph{Supervision Module} As described in the previous section, the supervision module incorporates the relation KBs and resolution data programming rules as seed self-supervision, as well as reasoning rules such as resolution transitivity for joint inference.  Note that these self-supervision rules can be noisy, but for simplicity we still treat them as hard constraints. Deep probabilistic logic offers a principled way to soften them and model their noisiness, which can be investigated in future work. 

% Learning obj/algo
%{\bf TODO: tweak algo to show rel/arg var EM together}

\paragraph{Learning} The prediction and supervision modules define a joint probabilistic distribution
% \vspace{-3pt}
\[P(K,Y|X)\propto \prod_{v\in K}~\Phi_{v}(X, Y)\cdot\prod_i~\Psi(X_i, Y_i)\] %\vspace{-3pt}
Here, $K$ represents the self-supervision and $(X_i, Y_i)$ the input-output pairs of relation detection and argument resolution. $\Phi, \Psi$ are the supervision and prediction modules, respectively. 
Learning is done via variational EM. In the E-step, we compute a variational approximation $q(Y)\propto P(Y|K,X)$ using loopy belief propagation, based on current $\Phi,\Psi$. In the M-step, we treat $q(Y)$ as the probabilistic labels and refine parameters of $\Phi,\Psi$. 
As aforementioned, we treat the self-supervision in $\Phi$ as hard constraints, so the M-step simplifies to fine-tuning the  transformer-based models for relation detection and argument resolution, treating $q(Y)$ as probabilistic labels.

\paragraph{Inference} After learning, given a test document and candidate entities and mentions, it is straightforward to run the neural modules for relation detection and argument resolution. Additionally, we would incorporate joint inference for argument resolution as in self-supervision (e.g., transitivity) using loopy belief propagation. 

\section{Experiments}

\begin{table}[!t]
\small
\centering
\begin{tabular}{@{}lrrr@{}}
\toprule
 & CKB & ~~~~\ckbhard & DocRED \\ \midrule
\textbf{Documents} & 430 & 391~ & 1000 \\
%\textbf{Unique facts} & 1566 & 325~ & 11790 \\
\textbf{Relations} & 1904 & 332~ & 12323 \\
%\textbf{Negative} & 15840 & 11790~ & 186072 \\
\textbf{Candidates} & 17744 & 12122~ & 198395 \\
\multicolumn{4}{l}{\hspace{-0.62em}\textbf{Document-level statistics (mean)}}  \\
~~~~Words & 5480.1 & 5576.2~ & 200.7 \\
~~~~Sentences & 170.6 & 173.6~ & 8.1 \\
~~~~Paragraphs & 38.9 & 39.3~ & 1.0 \\
\multicolumn{4}{l}{\textbf{\hspace{-0.62em}Minimal span for relations (mean)}}  \\
~~~~Words & 161.8 & 717.7~ & 30.9 \\
~~~~Sentences & 5.7 & 22.6~ & 2.2 \\
~~~~Paragraphs & 1.7 & 4.8~ & 1.0 \\ \bottomrule
\end{tabular}
\caption{Comparison of statistics among CKB, \ckbhard, and DocRED (validation; test annotations are not publicly available). DocRED comprises short Wikipedia introduction sections, whereas CKB features full-text articles. CKB also features relations spanning much longer text, especially in the \ckbhard subset.
}
\label{tab:test-data-statistics}
\end{table}

% \begin{table}[!ht]
% \small
% \centering
% \begin{tabular}{@{}lrr@{}}
% \toprule
%  & Train & Dev \\ \midrule
% \textbf{Documents} & 5563 & 2400 \\
% %\textbf{Unique facts} & 274 & 357 \\
% \textbf{Positive examples} & 5019 & 5536 \\
% \textbf{Negative examples} & 12367 & 6582 \\ \bottomrule
% \end{tabular}
% \caption{Distant supervision for relation detection.}
% \label{tab:train-dev-data-for-relation-dectection}
% \end{table}

% overview
In this section, we study how our modular self-supervision approach performs in document-level relation extraction. A popular dataset is DocRED~\cite{yao-etal-2019-docred}, which features Wikipedia introduction sections and general-domain relations. However, upon close inspection, DocRED does not have many truly long-range relation instances. As \autoref{tab:test-data-statistics} shows, each DocRED document contains only eight sentences in average, most within a single paragraph. About half of relations (49\%) can be extracted from a single sentence, per evidence annotation. Consequently, there is very little room to explore the more challenging scenario where relations span multiple paragraphs in large text spans. 
In fact,
\citet{huang-etal-2021-three} finds that over 95\% instances in DocRED require no more than three sentences as supporting evidence, and 87\% requires no more than two sentences. \citet{ye-etal-2020-coreferential} shows that a simple BERT-based system (a special case of our approach with just local relation detection) yields 60.06\% F1 on DocRED test (Table 3 in their paper), very close to the state-of-the-art results of 62.76\% by GAIN~\cite{zeng-etal-2020-double}.

We thus focus on biomedical machine reading, where there is a pressing need for comprehensive extraction of the latest findings from full-text articles, and cross-paragraph relation mentions are prevalent. Following \citet{peng-etal-2017-cross,jia-etal-2019-document}, we consider the problem of extracting precision oncology knowledge from PubMed Central full-text articles, which is critical for molecular tumor boards and other precision health applications. Concretely, the goal is to extract drug-gene-mutation relations as shown in \autoref{fig:motivating-example}: given a drug, gene, mutation, and document in which they are mentioned, determine whether the document asserts that the mutation in the gene affects response to the drug. 

\subsection{Datasets}

\begin{table}[!t]
\small
\centering
\begin{tabular}{@{}lrrr@{}}
\toprule
 & Documents & Positive ex. & Negative ex. \\ \midrule
Train & 5563 & 5019 & 12367 \\
Dev & 2400 & 5539 & 6582 \\ \bottomrule
\end{tabular}
\caption{Distant supervision for relation detection.}
\label{tab:train-dev-data-for-relation-dectection}
\end{table}

\paragraph{Self-Supervision} For training and development, we use unlabeled documents from the PubMed Central Open Access Subset (PMC-OA)\footnote{\url{www.ncbi.nlm.nih.gov/pmc/}}. For relation detection, we derive distant supervision from three knowledge bases (KBs) with manually-curated drug-gene-mutation relations: CIVIC\footnote{\url{civicdb.org/home}}, GDKD~\cite{dienstmann2015database}, OncoKB~\cite{chakravarty2017oncokb}.
We randomly split the generated examples into training and development sets and ensure no overlap of documents. 
\autoref{tab:train-dev-data-for-relation-dectection} summarizes their statistics.
For argument resolution, we use the global reasoning rules such as transitivity, as well as data programming rules capturing two anaphoric phenomena: identical mentions, apposition. 

%{\bf For argument resolution, what's train/dev stats?}

\paragraph{Evaluation} Following \citet{jia-etal-2019-document}, we 
use \ckbcore from the Clinical Knowledgebase (CKB;~\citealt{patterson2016clinical})\footnote{\url{ckbhome.jax.org}} as our gold-standard test set. 
\ckb contains high-quality document-level annotations of drug-gene-mutation interactions, which are manually curated from PubMed articles by The Jackson Laboratory (JAX), an NCI-designated cancer center. 
\ckb has minimal overlap with the three KBs used in training and development. To avoid contamination, we remove CKB entries whose documents are used in our training and development. See \autoref{tab:test-data-statistics} for statistics. 
Note that compared to the version used in \citet{jia-etal-2019-document}, the latest dataset (accessed in Oct. 2020) contains substantially more relations from recent findings.
For about one fifth of annotated relations (17.4\%), the key entities such as drug and mutation never co-occur in the same paragraph. These relations are out of scope in \citet{jia-etal-2019-document}. 
We denote this subset as \ckbhard, which comprises particularly challenging instances requiring cross-paragraph discourse modeling. 
%While \ckbhard contains 332/1904 (17.4\%) positive examples of the entire test set, its negative example rate is overwhelming.
%To our knowledge, CKB CORE\texttrademark~is the only manually curated data that contains a significant portion of cross-paragraph relational facts.

\subsection{Systems}

% sarthak: Table -> \small . Not sure if there is a reason you want to make it normal size. Please feel free to change back.
\begin{table*}[!t]
\centering
\small
\begin{tabular}{@{}C{2cm}lR{2.2cm}R{2.5cm}R{2.5cm}@{}}
\toprule
Test Set & System & Precision (\%) & Recall (\%) & F1 (\%)~~ \\ \midrule
\multirow{4}{*}{CKB} & \allpos & 10.7 {\textcolor{white}{\footnotesize $\pm$0.0}} & 100.0 {\textcolor{white}{\footnotesize $\pm$0.0}} & 19.4 {\textcolor{white}{\footnotesize $\pm$0.0}}~~ \\  
& \multiscale~\cite{jia-etal-2019-document} & 21.2 {\footnotesize$\pm$0.5} & 59.3 {\footnotesize$\pm$0.2} & 31.2 {\footnotesize$\pm$0.5}~~ \\
 & \gain~\cite{zeng-etal-2020-double} & 42.3 {\footnotesize$\pm$2.0} & 33.5 {\footnotesize$\pm$1.5} & 37.3 {\footnotesize$\pm$0.2}~~ \\
 & \modular (Ours) & 49.9 {\footnotesize$\pm$0.3} & 71.5 {\footnotesize$\pm$0.3} & {\bf 58.8} {\footnotesize$\pm$0.1}~~ \\ \midrule
\multirow{4}{*}{~~~\ckbhard} & \allpos & 2.7 {\textcolor{white}{\footnotesize $\pm$0.0}} & 100.0 {\textcolor{white}{\footnotesize $\pm$0.0}} & 5.3 {\textcolor{white}{\footnotesize $\pm$0.0}}~~ \\  
 & \multiscale~\cite{jia-etal-2019-document} & 4.7 {\footnotesize$\pm$0.2} & 42.8 {\footnotesize$\pm$0.9} & 8.5 {\footnotesize$\pm$0.3}~~ \\
 & \gain~\cite{zeng-etal-2020-double} & 7.9 {\footnotesize$\pm$0.4} & 13.7 {\footnotesize$\pm$1.4} & 10.0 {\footnotesize$\pm$0.4}~~ \\
 & \modular (Ours) & 22.6 {\footnotesize$\pm$0.1} & 46.1 {\footnotesize$\pm$0.4} & {\bf 30.3} {\footnotesize$\pm$0.1}~~ \\ \bottomrule
\end{tabular}
\caption{Comparison of test results on \ckb and \ckbhard. Relations in \ckbhard are particularly challenging as key entity pairs such as drug and mutation never co-occur in a paragraph. All systems were trained using the same three KBs for distant supervision, with no overlap with \ckb. We report results from three random runs.}
\label{tab:main-results}
\end{table*}

We implemented our modular self-supervision method (\modular) using PyTorch~\cite{pytorch}. 
We conducted variational EM for eight iterations, which appear to work well in preliminary experiments. 
In the M-step, we incorporate early stopping to identify the best checkpoint based on the development performance for fine-tuning the relation detection and argument resolution neural modules. 
We initialized the encoding layers in both modules with PubMedBERT~\cite{pubmedbert}, which has been pretrained from scratch on PubMed articles and demonstrated superior performance in a wide range of biomedical NLP applications. 

We follow \citet{wang-poon-2018-deep,jia-etal-2019-document} to conduct standard data preprocessing and entity linking.
We used the AdamW optimizer~\cite{loshchilov2017decoupled}. 
%See the Appendix for details. %ed hyperparameter setttings.
For training, we set the mini-batch size to 32 and the learning rate 5e-5 with 100 warm-up steps and 0.01 weight decay. The drop-out rate is 0.1 for transformer-based encoders, and 0.5 for other layers. The hidden size is 765 for transformer-based encoders, and 128 for all other feed-forward networks. We generate checkpoints at every 4096 steps. Three random seeds are tried in our experiments: [7, 12, 17].
% tjn: This is an excellent description.

For self-supervised relation detection, following \citet{jia-etal-2019-document}, we further decompose it into classifying drug-mutation relations and then augmenting them with high-precision gene-mutation associations. 
As stated in Section \ref{sec:modular}, at training time, only named entities such as drug, gene, mutation are considered, whereas at inference time, in principle any co-occurring noun phrases within a paragraph would be considered (see \autoref{fig:formulation}, bottom paragraph). In practice, however, this would incur too much computation, most of which wasted on irrelevant candidates. Therefore, we employ the following heuristics to leverage argument resolution results for filtering candidates: In argument resolution, we focus on resolving candidate mentions with drugs, genes, mutations. We also stipulate that a candidate mention must contain within it some relevant biomedical entity mentions (e.g., cell lines, genes, etc., as in ``MEK inhibitors'' that contains gene reference ``MEK''). 
In relation detection, we only consider candidate mentions that are classified as resolving with entities among drugs, genes, mutations, based on current prediction module.

We compare \modular with the following baselines:
\allpos is a recall-friendly baseline that always predicts positive;
\multiscale~\cite{jia-etal-2019-document} is a state-of-the-art approach that combines local mention-level representations into an entity-level representation over the entire document;
\gain~\cite{zeng-etal-2020-double} is another state-of-the-art approach that constructs mention-level graphs and applies graph convolutional network~\cite{kipf2017semi} to model interdependencies among intra- and inter-sentence mentions, attaining top performance on DocRED.

For fair comparison, we replaced the encoders in \multiscale and \gain with the state-of-the-art PubMedBERT as in our approach, which helped improve the performance (Appendix \ref{sec:Jia-pubmedbert}). 
The original \multiscale encodes each paragraph separately using LSTM, so it's straightforward to replace that with PubMedBERT. \gain, on the other hand, encodes the entire input text all at once. This is feasible in DocRED, where each ``document'' is actually a Wikipedia introduction section, thus more like a paragraph (average only eight sentences long). But it doesn't work in \ckb, where each document is a full-text article. Even the minimal text span covering given entities is often too long to encode using a transformer. Therefore, we ran the encoder on individual paragraphs. 
Note that the original version of \multiscale can't make prediction for any instances where key entity pairs such as drug and mutation never co-occur in a paragraph. We implemented a natural extension that would generate local mention-level representations even for singleton mentions (i.e., only one relevant entity shows up in a paragraph). 

%{\bf Encode individual paragraphs - then ...}
%{\bf Did we confirm that PubMedBERT perform better for them?}
%{\bf How did we deal w. long text spans for multiscale and gain?}

\subsection{Main Results}
\label{sec:results}

\Cref{tab:main-results} compares various approaches for document-level relation extraction on \ckb.
Our modular self-supervision approach (\modular) substantially outperforms all other methods, gaining more than 20 absolute F1 points compared to prior state of the art such as multiscale learning and graph neural networks. 
This demonstrates the superiority in leveraging less noise-prone modular self-supervision as well as fine-grained discourse modeling in argument resolution. 
Note that the results for \multiscale are different than that in \citet{jia-etal-2019-document} as we used the latest \ckb which contains considerably more cross-paragraph relations. 

Compared to multiscale learning (\multiscale), the graph-neural-network approach (\gain) attains significantly better precision, as it incorporates more elaborate graph-based reasoning among entities across sentences. However, this comes with substantial expense at recall. Our approach outperforms both substantially in precision and recall. 

On the most challenging subset \ckbhard, the contrast is particularly pronounced, as all other systems could only attain single-digit precision. The graph-neural-network approach also suffers heavily in recall. Our approach attains much better precision and recall, and more than triples the F1.

\begin{table}[!t]
\centering
\small
\begin{tabular}{@{}lrrr@{}}
\toprule
 & Prec. & Recall & F1 \\ \midrule
\modular (Ours) & 22.6 & 46.1 & {\bf 30.3} \\
\makecell[bl]{$\triangleright$~Replace relation detection\\~~~with \allpos} & 13.7 & 66.1 & 22.7 \\ 
\makecell[bl]{$\triangleright$~Replace argument resolution\\~~~with Multi-Sieve Pass} & 11.0 & 42.8 & 17.5 \\
\makecell[bl]{$\triangleright$~Replace argument resolution\\~~~with SpanBERT Coref} & 60.2 & 1.5 & 2.9 \\
% \makecell[bl]{$\triangleright$~Remove noisy data filter}& 26.1 & 38.8 & 26.4 \\ 
\bottomrule
\end{tabular}
\caption{Ablation study on \ckbhard.}
\label{tab:ablation-study}
\end{table}

% \vspace{-2pt}
\subsection{Ablation Study}

%Compared to prior state of the art such as \multiscale and \gain, our \modular approach decomposes end-to-end extraction into relation detection and argument resolution modules. 
To understand the impact of our modular self-supervision, we conducted an additional ablation study on \ckbhard. See \Cref{tab:ablation-study} for results. 

To assess the limitation of our current argument resolution module, we replace self-supervised relation detection with a baseline that always predicts positive for candidate tuples whose components have been resolved with some drug, gene, mutation entities. This yields a maximum recall of 66.1\%, which means that about a third of the especially hard cases of cross-paragraph relations are still out of reach for our method. In some cases, this is because the only hint at the relation resides in figures or appendix, which are currently not in scope for extraction. In other cases, the argument resolution fails to make correct resolution with the corresponding entities. We leave further investigation and improvement to future work.
With self-supervised relation detection, our full model improves both F1 and precision for the end extraction on \ckbhard. 
%Note that with cross-paragraph relations, the number of candidates is two orders of magnitude larger than the number of true relation instances (~12,000 vs. 332), therefore, the precision gain means that 

Next, we investigate the impact of our self-supervision for argument resolution by replacing it with state-of-the-art coreference systems: Multi-Sieve Pass~\cite{raghunathan-etal-2010-multi,lee-etal-2011-stanfords},
and SpanBERT Coreference~\cite{joshi-etal-2020-spanbert}. 
Multi-Sieve Pass is a rule-based system that incorporates a series of resolution rules with increasing recall but lower precision. SpanBERT Coreference is a state-of-the-art transformer-based system fine-tuned on OntoNotes~\cite{ontonotes}, an annotated corpus with diverse text. 
Both result in significant performance drop. 
Using Multi-Sieve results in substantial drop in precision, indicating that coreference heuristics suitable for  general domains are less effective in biomedicine. 
SpanBERT, on the other hand, suffers catastrophic drop in recall. This suggests that argument resolution for document-level relation extraction may involve more general anaphoric phenomena such as $\tt ISA$ and $\tt PartOf$, which are out of scope in standard coreference annotations. 
Remarkably, bootstrapping from the simple data programming rules of identical mentions and apposition, our self-supervised module is able to perform much better argument resolution than these state-of-the-art systems for document-level relation extraction. 

% {\bf What's noisy data filter?}
% Lastly, we remove the noisy data filter when training the relation component. While precision slightly increases, both F1 and recall drop significantly.

\begin{table}[!ht]
\centering
\small
\begin{tabular}{@{}clccc@{}}
\toprule
Test Data & Pre-trained Encoder & Prec. & Recall & F1 \\ \midrule
\multirow{2}{*}{\ckb} & PubMedBERT & 49.9 & 71.5 & {\bf 58.8} \\
 & BERT & 49.5 & 68.6 & 57.5 \\ \midrule
\multirow{2}{*}{~~\ckbhard} & PubMedBERT & 22.6 & 46.1 & {\bf 30.3} \\
 & BERT & 21.8 & 39.2 & 28.1 \\ \bottomrule
\end{tabular}
\caption{Domain-specific pretraining improves test performance over general-domain pretraining.}
\label{tab:bert-comp}
\end{table}

%{\bf Replace SpanBERT w. BERT; add overall ckb results as well}

% \vspace{-2pt}
Given that our evaluation is in the biomedical domain, it is natural to initialize our self-supervised neural modules with PubMedBERT~\cite{pubmedbert}. \autoref{tab:bert-comp} shows that this is indeed advantageous, with domain-specific pretraining attains significant gain over general-domain pretraining.

\subsection{Discussion}

\paragraph{Interpretability} We envision that machine reading is used not as standalone automation, but as assisted curation to help expert curators attain significant speed-up \cite{peng-etal-2017-cross}. For extraction within short text spans, human experts can validate the results by simply reading through the provenance text. For document-level relation extraction, as in \citet{jia-etal-2019-document}, this can be challenging, as the intervening text span is long and validation may require a significant amount of reading that is not much faster than curation from scratch. Our modular approach not only enables us to tackle the harder cases of cross-paragraph relations, but also yields a natural explanation for an extraction result with the local relation and chains of argument resolution, all of which can be quickly validated by curators. We leave studying the impact on assisted curation to future work.

\noindent\textbf{Error Analysis} 
We focus our error analysis on \ckbhard, which is particularly challenging. An immediate opportunity lies in significant recall miss by argument resolution. As shown in the prior subsection, the maximum recall for our current argument resolution module is 66\%. From preliminary sample analysis, there are three main types of errors. Some relation instances are hinted in figures, tables, or supplements, which are beyond the current scope of extraction. In other cases, the relation statement is vague and scattered, and relation detection requires inference piecing together multiple evidences. However, the bulk of recall errors simply stem from argument resolution failures. Likewise, we found that in the majority of precision errors, relation detection appears to make the right call, but argument resolution is mistaken.

\begin{figure}[!ht]
    \centering
    \includegraphics[width=0.43\textwidth]{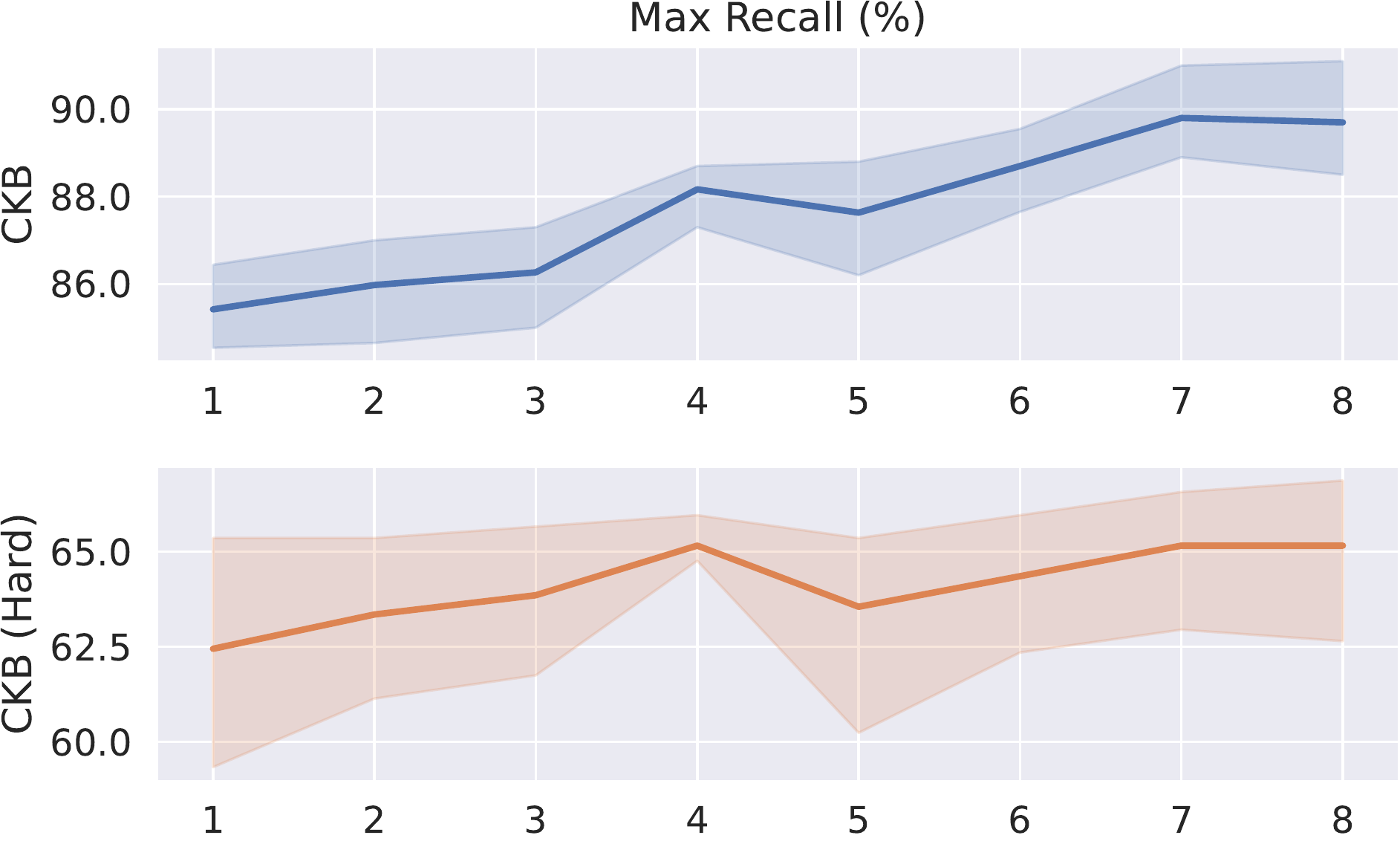}
    \caption{Argument resolution improves during learning as global reasoning rules augment seed self-supervision to raise maximum recall.}
    \label{fig:resolution-component-learning}
\end{figure}

\paragraph{Variational EM} 
One direction to improve argument resolution is to augment the self-supervision used in the resolution module. \Cref{fig:resolution-component-learning} shows that argument resolution did improve during learning, thanks to the global reasoning rules, at least in terms of expanding maximum recall. 
However, we notice that our current mention filtering rules may be overly strict, which limit the room for growth in recall. Additionally, we treat the reasoning rules such as transitivity as hard constraints, whereas in practice they can be noisy. (E.g., qualifiers like ``some MAP2K1 mutations'' or negation are currently not considered in resolution.)

\section{Related Work}

% Doc-level RE 
\paragraph{Document-Level Relation Extraction}
Due to the significant challenges in modeling long text spans and obtaining high-quality supervision signals, document-level relation extraction has been relatively underexplored~\cite{surdeanu2014overview}. 
Prior work often focuses on simple extensions of sentence-level extraction (e.g., by incorporating coreference annotations or considering special cases when document-level relations reduces to sentence-level attribute classification)~\cite{wick-etal-2006-learning,gerber-chai-2010-beyond,swampillai-stevenson-2011-extracting,yoshikawa2011coreference,koch-etal-2014-type,yang-mitchell-2016-joint}. 
Recently, cross-sentence relation extraction has seen increasing interest~\cite{li2016biocreative,quirk-poon-2017-distant,peng-etal-2017-cross,verga-etal-2018-simultaneously,christopoulou-etal-2019-connecting,wu2019renet,yao-etal-2019-docred}, but most efforts are still limited to short text spans, such as consecutive sentences or abstracts. 
A notable exception is \citet{jia-etal-2019-document}, which considers full-text articles that comprise hundreds of sentences. However, they still model local text units in isolation and can't effectively handle relations whose arguments never co-occur in a paragraph. 
In contrast, we provide the first attempt to systematically explore cross-paragraph relation extraction. 
Most prior work focuses on binary relations. We instead follow \citet{peng-etal-2017-cross,jia-etal-2019-document} to study general $n$-ary relation extraction, using precision oncology treatment as a case study. 

% coref / discourse
\paragraph{Discourse Modeling} Given the focus of standard information extraction on short text spans, discourse modeling has not featured prominently in prior work. An exception is coreference resolution, though the focus tends to be improving sentence-level extraction, as in \citet{koch-etal-2014-type}. Here, we show that document-level relation extraction often requires modeling more general anaphoric phenomena. As discussed in the experiment section, many remaining errors lie in argument resolution, which offers an exciting opportunity to study discourse modeling for an important end application.

% Self-supervision
\paragraph{Self-Supervision} Task-specific self-supervision alleviates the annotation bottleneck by leveraging freely available domain knowledge (as in distant supervision~\cite{craven1999constructing,mintz-etal-2009-distant}) and expert-derived labeling rules (as in data programming~\cite{snorkel}). Unfortunately, such self-supervision becomes extremely noisy when applied to full-text documents, prompting many prior efforts to focus on short text spans~\cite{quirk-poon-2017-distant,peng-etal-2017-cross,verga-etal-2018-simultaneously,yao-etal-2019-docred}. We instead decompose end-to-end document-level extraction into relation detection and argument resolution modules, for each of which we leverage modular self-supervision that is much less error-prone. 

% Neural-Symbolic AI
\paragraph{Neural-Symbolic NLP} In the past few decades, the dominant paradigm in NLP has swung from logical approaches (rule-based or relational systems) to statistical and neural approaches. However,  given the prevalence of linguistic structures and domain knowledge, there has been increasing interest for synergizing the contrasting paradigms to improve inference and learning. 
Neural logic programming replaces logical operators with neural representations to leverage domain-specific constraints with end-to-end differentiable learning~\cite{rocktaschel&riedel16}. Similarly, modular neural networks integrate component neural learning along a structured scaffold (e.g., syntactic parse of a sentence for visual question-answering)~\cite{andreas2016neural}. 
On the other hand, deep probabilistic logic~\cite{wang-poon-2018-deep} combines probabilistic logic with neural networks to incorporate diverse self-supervision for deep learning. We take inspiration from modular neural networks and neural logic programming, and use deep probabilistic logic to combine relation detection and argument resolution using global reasoning rules for document-level relation extraction. 
\section{Conclusion}

We propose to decompose document-level relation extraction into local relation detection and global argument resolution, and apply modular self-supervision and discourse modeling using deep probabilistic logic. On the challenging problem of biomedical machine reading, where cross-paragraph relations are prevalent, our approach substantially outperforms prior state of the art such as multiscale learning and graph neural networks, gaining over 20 absolute F1 points. 
Future directions include: improving discourse modeling for argument resolution; studying the impact on assisted curation; applications to other domains.
\section*{Acknowledgements}
%We give warm thanks to the Project Hanover team %especially Rajesh Rao and Michael Lucas We also thank Michel Galley and the anonymous reviewers for their helpful comments.
We give warm thanks to the Project Hanover team, Michel Galley, and the anonymous reviewers for their helpful comments.

%\input{sections/problem_formulation}
%\input{sections/model}
%\input{sections/learning}
%\input{sections/inference}
%\input{sections/experiments}
%\input{sections/related_work}

\bibliography{anthology,custom}
\bibliographystyle{acl_natbib}

\newpage
\appendix

\section{Neural Architectures}
\label{sec:neural-architecture}

\begin{figure}[!ht]
    \centering
    \includegraphics[width=0.4\textwidth]{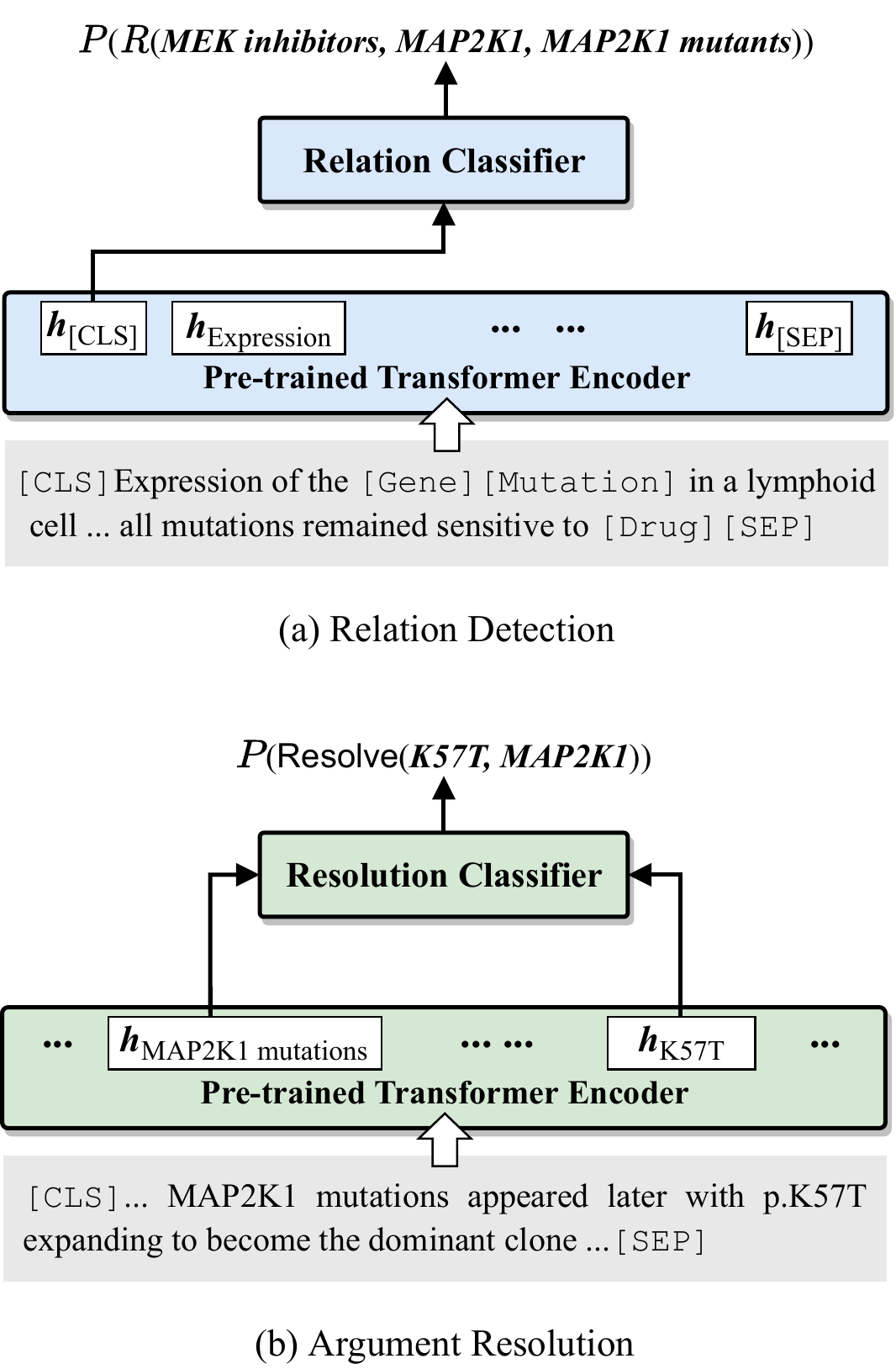}
      \caption{Neural architectures for relation detection and argument resolution.}
    \label{fig:model}
\end{figure}

\begin{table}[!t]
\small
\center
\begin{tabular}{|l|c|ccc|}
\hline
System & AUC & Prec. & Recall & F1 \\ \hline
\hline
\textbf{Base versions} &  &  &  &  \\
\textsc{SentLevel} &  &  &  &  \\
\quad\citet{jia-etal-2019-document} & 22.4 & 39.3 & 34.7 & 36.9 \\
\quad Our reproduction & {\color[HTML]{CB0000} 22.4} & {\color[HTML]{CB0000} 38.9} & {\color[HTML]{CB0000} 35.5} & {\color[HTML]{CB0000} 37.1} \\
\qquad w. PubMedBERT & {\color[HTML]{3166FF} 24.4} & {\color[HTML]{3166FF} 43.6} & {\color[HTML]{3166FF} 35.1} & {\color[HTML]{3166FF} 38.9} \\ \hline
\textsc{ParaLevel} &  &  &  &  \\
\quad\citet{jia-etal-2019-document} & 33.1 & 36.5 & 44.6 & 40.1 \\
\quad Our reproduction & {\color[HTML]{CB0000} 32.8} & {\color[HTML]{CB0000} 35.6} & {\color[HTML]{CB0000} 44.3} & {\color[HTML]{CB0000} 39.5} \\
\qquad w. PubMedBERT & {\color[HTML]{3166FF} 33.2} & {\color[HTML]{3166FF} 49.4} & {\color[HTML]{3166FF} 39.5} & {\color[HTML]{3166FF} 43.9} \\ \hline
\textsc{DocLevel} &  &  &  &  \\
\quad\citet{jia-etal-2019-document} & 36.7 & 45.4 & 38.5 & 41.7 \\
\quad Our reproduction & {\color[HTML]{CB0000} 37.0} & {\color[HTML]{CB0000} 43.3} & {\color[HTML]{CB0000} 41.9} & {\color[HTML]{CB0000} 42.6} \\
\qquad w. PubMedBERT & {\color[HTML]{3166FF} 34.9} & {\color[HTML]{3166FF} 47.7} & {\color[HTML]{3166FF} 41.6} & {\color[HTML]{3166FF} 44.5} \\ \hline
\textsc{MultiScale} &  &  &  &  \\
\quad\citet{jia-etal-2019-document} & 37.3 & 41.8 & 43.4 & 42.5 \\
\quad Our reproduction & {\color[HTML]{CB0000} 36.9} & {\color[HTML]{CB0000} 38.5} & {\color[HTML]{CB0000} 46.2} & {\color[HTML]{CB0000} 42.0} \\
\qquad w. PubMedBERT & {\color[HTML]{3166FF} 35.9} & {\color[HTML]{3166FF} 43.4} & {\color[HTML]{3166FF} 46.3} & {\color[HTML]{3166FF} 44.8} \\ \hline
\hline
\textbf{+ Noisy-Or} &  &  &  &  \\
\textsc{SentLevel} &  &  &  &  \\
\quad\citet{jia-etal-2019-document} & 25.3 & 39.3 & 35.3 & 37.2 \\
\quad Our reproduction & {\color[HTML]{CB0000} 25.3} & {\color[HTML]{CB0000} 38.4} & {\color[HTML]{CB0000} 35.9} & {\color[HTML]{CB0000} 37.1} \\
\qquad w. PubMedBERT & {\color[HTML]{3166FF} 26.0} & {\color[HTML]{3166FF} 43.6} & {\color[HTML]{3166FF} 35.2} & {\color[HTML]{3166FF} 39.0} \\ \hline
\textsc{ParaLevel} &  &  &  &  \\
\quad\citet{jia-etal-2019-document} & 35.6 & 44.3 & 40.6 & 42.4 \\
\quad Our reproduction & {\color[HTML]{CB0000} 35.5} & {\color[HTML]{CB0000} 45.5} & {\color[HTML]{CB0000} 39.2} & {\color[HTML]{CB0000} 42.1} \\
\qquad w. PubMedBERT & {\color[HTML]{3166FF} 37.3} & {\color[HTML]{3166FF} 51.6} & {\color[HTML]{3166FF} 38.8} & {\color[HTML]{3166FF} 44.2} \\ \hline
\textsc{DocLevel} &  &  &  &  \\
\quad\citet{jia-etal-2019-document} & 36.7 & 45.4 & 38.5 & 41.7 \\
\quad Our reproduction & {\color[HTML]{CB0000} 37.0} & {\color[HTML]{CB0000} 43.3} & {\color[HTML]{CB0000} 42.0} & {\color[HTML]{CB0000} 42.6} \\
\qquad w. PubMedBERT & {\color[HTML]{3166FF} 34.9} & {\color[HTML]{3166FF} 47.7} & {\color[HTML]{3166FF} 41.6} & {\color[HTML]{3166FF} 44.5} \\ \hline
\textsc{MultiScale} &  &  &  &  \\
\quad\citet{jia-etal-2019-document} & 39.7 & 48.1 & 38.9 & 43.0 \\
\quad Our reproduction & {\color[HTML]{CB0000} 39.6} & {\color[HTML]{CB0000} 48.7} & {\color[HTML]{CB0000} 37.8} & {\color[HTML]{CB0000} 42.6} \\
\qquad w. PubMedBERT & {\color[HTML]{3166FF} 39.5} & {\color[HTML]{3166FF} 46.8} & {\color[HTML]{3166FF} 43.1} & {\color[HTML]{3166FF} 44.9} \\ \hline
\end{tabular}
\caption{Comparison on the original CKB test set: reported results in \citet{jia-etal-2019-document}, results from our reproduction,
and results after replacing LSTM with PubMedBERT as the encoder.}
\label{tab:pubmedbert-vs-lstm}
\end{table}

\Cref{fig:model} shows the neural architectures for relation detection and argument resolution.
For relation detection, we input a paragraph to a transformer-based encoder, with mentions dummified.
The hidden state $\bm{h}_\texttt{[CLS]}$ in the last layer is then passed to a simple feed-forward classifier defined as below:
\begin{equation*}
    \text{classifier}(\bm{h})=\text{sigmoid}(\text{FFNN}_1(\bm{h}))\\
\end{equation*}
where FFNN$_1$ is a two-layer feed-forward network using ReLU as activation functions. 

For argument resolution, given a candidate mention pair ($m_i,m_j$) and their context,
we first compute their contextual representations using a transformer-based encoder.
If a mention span contains multiple tokens,
we use average pooling to combine their contextual representations (hidden states in the last layer). 
The pair of mention representations ($\bm{h}_{m_i}, \bm{h}_{m_j}$) are then passed to a classifier defined as below:
\begin{equation*}
    \text{classifier}(\bm{h}_{m_i},\bm{h}_{m_j})=\text{sigmoid}(s(\bm{h}_{m_i}, \bm{h}_{m_j}))\\
\end{equation*}
where $s(\bm{x},\bm{y})$ is a scoring function similar to those used in \citet{lee-etal-2018-higher,joshi-etal-2020-spanbert}:
\begin{gather*}
    s(\bm{x},\bm{y})=s_m(\bm{x})+s_m(\bm{y})+s_c(\bm{x},\bm{y})\\
    s_m(\bm{x})=\text{FFNN}_2(\bm{x})\\
    s_c(\bm{x},\bm{y})=\text{FFNN}_3([\bm{x}, \bm{y}, \bm{x}\circ\bm{y}])
\end{gather*}
where $\circ$ denotes element-wise multiplication.
FFNN$_2$ and FFNN$_3$ are two-layer feed-forward networks using ReLU as activation functions.

\section{Multiscale with PubMedBERT}
\label{sec:Jia-pubmedbert}

Replacing LSTM with PubMedBERT \cite{pubmedbert} as the encoder generally leads to comparable or better performance by the \multiscale system \cite{jia-etal-2019-document}. \Cref{tab:pubmedbert-vs-lstm} shows the results on the original CKB test set as used in \citet{jia-etal-2019-document}. Note that these results shouldn't be compared with the main results in Table \ref{tab:main-results}, as the latter are obtained on the latest \ckb with considerably more cross-paragraph relations.

\end{document}